\documentclass[11pt,a4paper]{article}
\usepackage[hyperref]{emnlp2020}
\usepackage{times}
\usepackage{latexsym}
\usepackage{booktabs}
\usepackage{array}
\usepackage{colortbl}
\usepackage{graphicx}
\usepackage{amsmath}
\usepackage{multirow}
\graphicspath{ {./images/} }

\usepackage{cleveref}[2012/02/15]
\usepackage{microtype}

\aclfinalcopy 

\crefformat{footnote}{#2\footnotemark[#1]#3}

\title{HINT3: Raising the bar for Intent Detection in the Wild}

\author{Gaurav Arora \\
  Jio Haptik \\
  \texttt{gaurav@haptik.ai} \\\And
  Chirag Jain \\
  Jio Haptik \\
  \texttt{chirag.jain@haptik.ai} \\\AND
  Manas Chaturvedi \\
  Jio Haptik \\
  \texttt{manas.chaturvedi@haptik.ai} \\\And
  Krupal Modi \\
  Jio Haptik \\
  \texttt{krupal@haptik.ai} \\}

\date{}

\begin{document}
\maketitle
\begin{abstract}
Intent Detection systems in the real world are exposed to complexities of imbalanced datasets containing varying perception of intent, unintended correlations and domain-specific aberrations. To facilitate benchmarking which can reflect near real-world scenarios, we introduce 3 new datasets created from live chatbots in diverse domains. Unlike most existing datasets that are crowdsourced, our datasets contain real user queries received by the chatbots and facilitates penalising unwanted correlations grasped during the training process. We evaluate 4 NLU platforms and a BERT based classifier and find that performance saturates at inadequate levels on test sets because all systems latch on to unintended patterns in training data.
\end{abstract}

\section{Introduction}
Over the last few years, task-oriented dialogue systems have gained increasing traction for applications like personal assistants, automated customer support agents, etc. This has led to the availability of several commercialised and/or open conversational bot building platforms. Most popular systems today involve intent detection as a vital part of their Natural Language Understanding (NLU) pipeline. Recent advances in transfer learning \citep{howard-ruder-2018-universal,peters-etal-2018-deep,devlin-etal-2019-bert} has enabled systems that perform quite well on existing benchmarking datasets \citep{larson-etal-2019-evaluation,casanueva-etal-2020-efficient}.

Definitions of intent often vary across users, tasks and domains. Perception of intent could range from a generic abstraction such as “Ordering a product” to extreme granularity such as “Enquiring for a discount on a specific product if ordered using a specific card”. Additionally, factors such as imbalanced data distribution in the training set, assumptions during training data generation, diverse background of domain experts involved in defining the classes make this task more challenging. During inference, these systems may be deployed to users with diverse cultural backgrounds who might frame their queries differently even when communicating in the same language. Furthermore, during inference, apart from correctly identifying in-scope queries, the system is expected to accurately reject out-of-scope \citep{larson-etal-2019-evaluation} queries, adding on to the challenge.

Most existing datasets for intent detection are generated using crowdsourcing services. To accurately benchmark in real-world settings, we release 3 new single-domain datasets, each spanning multiple coarse and fine grain intents, with the test sets being drawn entirely from actual user queries on the live systems at scale instead of being crowdsourced. On these datasets, we find that the performance of existing systems saturates at unsatisfactory levels because they end up learning spurious patterns from the training dataset instead of generalising to the perceived meanings of intents.

\begin{table*}[h]
\centering
\resizebox{\textwidth}{!}{%
\begin{tabular}{cccccc}
\hline
\multirow{3}{*}{\textbf{Dataset}} & \multicolumn{2}{c}{\textbf{Example Intents}} & \multicolumn{3}{c}{\textbf{Example Queries}} \\ \cline{2-6} 
 & \multirow{2}{*}{\textbf{Type}} & \multirow{2}{*}{\textbf{Label}} & \textbf{Train} & \multicolumn{2}{c}{\textbf{Test}} \\ \cline{4-6} 
 &  &  & \textbf{} & \textbf{In-scope} & \textbf{Out-of-Scope} \\ \hline
\multirow{4}{*}{\begin{tabular}[c]{@{}c@{}}SOF\\ Mattress\end{tabular}} & \multirow{2}{*}{Generic} & \multirow{2}{*}{OFFERS} & \begin{tabular}[c]{@{}c@{}}What are the\\  available offers\end{tabular} & \multirow{2}{*}{Any other offers} & \multirow{3}{*}{\begin{tabular}[c]{@{}c@{}}If I order now do\\  i get 20\% discount\\  in lockdown period\end{tabular}} \\ \cline{4-4}
 &  &  & Give me some discount &  &  \\ \cline{2-5}
 & \multirow{2}{*}{Specific} & \multirow{2}{*}{\begin{tabular}[c]{@{}c@{}}100\_NIGHT\_\\ TRIAL\_OFFER\end{tabular}} & \begin{tabular}[c]{@{}c@{}}What is the\\  100-night offer\end{tabular} & \multirow{2}{*}{\begin{tabular}[c]{@{}c@{}}Free 100 days\\  trial\end{tabular}} &  \\ \cline{4-4} \cline{6-6} 
 &  &  & \begin{tabular}[c]{@{}c@{}}Trial offer on\\  customisation\end{tabular} &  & I need try \\ \hline
\multirow{3}{*}{Curekart} & \multirow{3}{*}{Generic} & \multirow{3}{*}{\begin{tabular}[c]{@{}c@{}}RECOMMEND\\ \_PRODUCT\end{tabular}} & \begin{tabular}[c]{@{}c@{}}Sir I want to fast \\ gain weight.\end{tabular} & \begin{tabular}[c]{@{}c@{}}I need Beginners\\  hair multivitamin\end{tabular} & \begin{tabular}[c]{@{}c@{}}Role of \\ electrolytes powder\end{tabular} \\ \cline{4-6} 
 &  &  & \begin{tabular}[c]{@{}c@{}}i'm beginner\\  in gym\end{tabular} & \begin{tabular}[c]{@{}c@{}}Which is the\\  best whey protein\end{tabular} & \begin{tabular}[c]{@{}c@{}}Is it help for\\  sperm count\end{tabular} \\ \cline{4-6} 
 &  &  & \begin{tabular}[c]{@{}c@{}}For biceps and\\  tricep muscle\\  growth supplements\end{tabular} & \begin{tabular}[c]{@{}c@{}}Can i get\\  rivamal 120 ml\\  at my home\end{tabular} & \begin{tabular}[c]{@{}c@{}}Can diabetic\\  patient have it\end{tabular} \\ \hline
\multirow{4}{*}{\begin{tabular}[c]{@{}c@{}}Power\\ play11\end{tabular}} & \multirow{4}{*}{Generic} & \multirow{4}{*}{\begin{tabular}[c]{@{}c@{}}CHAT\_WITH\_\\ AN\_AGENT\end{tabular}} & \begin{tabular}[c]{@{}c@{}}Connect with\\  agent\end{tabular} & \begin{tabular}[c]{@{}c@{}}chat with customer\\  service agent\end{tabular} & \multirow{2}{*}{\begin{tabular}[c]{@{}c@{}}Application is not\\  responding during\\  team joining\end{tabular}} \\ \cline{4-5}
 &  &  & \begin{tabular}[c]{@{}c@{}}My transaction\\  is incorrect\end{tabular} & \begin{tabular}[c]{@{}c@{}}PowerPlay11\\  rummy issue\end{tabular} &  \\ \cline{4-6} 
 &  &  & \begin{tabular}[c]{@{}c@{}}My sign up\\  bonus is incorrect\end{tabular} & \begin{tabular}[c]{@{}c@{}}Suddenly balance\\  gone\end{tabular} & \multirow{2}{*}{\begin{tabular}[c]{@{}c@{}}Why my current\\  basketball teams\\  being shown??\end{tabular}} \\ \cline{4-5}
 &  &  & \begin{tabular}[c]{@{}c@{}}Did not receive\\  my amount\end{tabular} & \begin{tabular}[c]{@{}c@{}}Why it is showing\\  wrong balance\end{tabular} &  \\ \hline
\end{tabular}%
}
\caption{Few examples of Intents and Queries in Train and Test set in HINT3 dataset}
\label{tab:example-queries-hint3}
\end{table*}

We evaluate 4 NLU platforms - Dialogflow\footnote{https://cloud.google.com/dialogflow}, LUIS\footnote{https://www.luis.ai/}, Rasa NLU\footnote{https://github.com/RasaHQ/rasa/},  Haptik\footnote{https://haptik.ai}\footnote{Access requests for signup on Haptik are processed via contact form at https://haptik.ai/contact-us/} and a BERT \citep{devlin-etal-2019-bert} based classifier on all 3 datasets and highlight gaps in language understanding. We further probe into queries where all the current systems fail and question the efficacy of the current approach of learning. Additionally, we repeat all our experiments on the subset of training data and show a performance drop in all the systems despite retaining relevant and sufficient utterances in the training subset. We've made our datasets and code freely accessible on GitHub to promote transparency and reproducibility\footnote{https://github.com/hellohaptik/HINT3}.

\section{Prior Work}

Despite intent detection being an important component of most dialogue systems, very few datasets have been collected from real users. Web Apps, Ask Ubuntu and Chatbot datasets from \citep{braun-etal-2017-evaluating} contain a limited number of intents (\textless 10), oversimplifying the task. More recent datasets like HWU64 from \citep{XLiu.etal:IWSDS2019} and CLINC150 from \citep{larson-etal-2019-evaluation} span a large number of intents in multiple domains but are generated using crowd sourcing services hence are limited in diversity in user expressions which arise from but not limited to domain specific presumptions, context from how and where the bot is made available, paraphrases emerging from cultural and ethnic diversity of user base, conversational slang, etc. Our work has some similarity with CLINC150, in that they also highlight the problem of out-of-scope intent detection and with BANKING77 from \citep{casanueva-etal-2020-efficient} that focuses on a single domain. However, all three - HWU64, CLINC150, BANKING77 offer relatively large and well balanced training set which might not be always feasible to collect for every new domain. For all datasets mentioned so far, recent works have reported a reasonably high performance (\textgreater 90\% average) for in-scope queries. Despite this, gaps in language understanding become apparent when such systems are deployed. Datasets introduced in this paper and further analysis of results attempts to recognise critical gaps in language understanding and calls for further research into more robust methods.

\begin{table}[]
\centering
\resizebox{\columnwidth}{!}{%
\begin{tabular}{llllll}
\hline
\multicolumn{1}{c}{\textbf{Dataset}} & \multicolumn{1}{c}{\textbf{\#Intent}} & \multicolumn{4}{c}{\textbf{\#Queries}} \\ \hline
 &  & \multicolumn{2}{c}{\textbf{Train}} & \multicolumn{2}{c}{\textbf{Test}} \\ 
 &  & \textbf{Full} & \textbf{Subset} & \textbf{in-scope} & \textbf{oos} \\ \hline
SOFMattress & 21 & 328 & 180 & 231 & 166 \\ 
Curekart & 28 & 600 & 413 & 452 & 539 \\ 
Powerplay11 & 59 & 471 & 261 & 275 & 708 \\ \hline
\end{tabular}%
}
\caption{Statistics of the 3 datasets in HINT3}
\label{tab:data-stats}
\end{table}

\begin{figure*}[h]
\includegraphics[width=8cm, height=4.20cm]{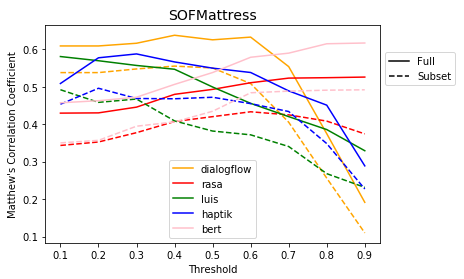}
\includegraphics[width=8cm, height=4.20cm]{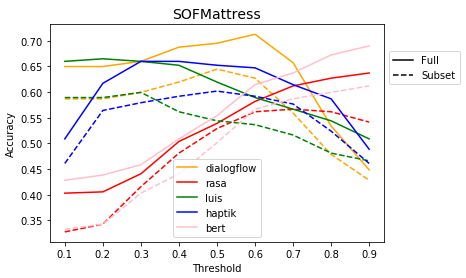}

\includegraphics[width=8cm, height=4.20cm]{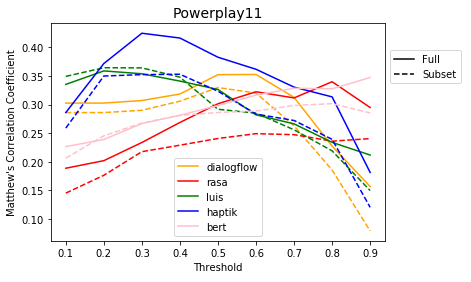}
\includegraphics[width=8cm, height=4.20cm]{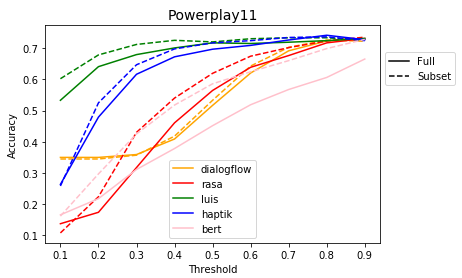}

\includegraphics[width=8cm, height=4.20cm]{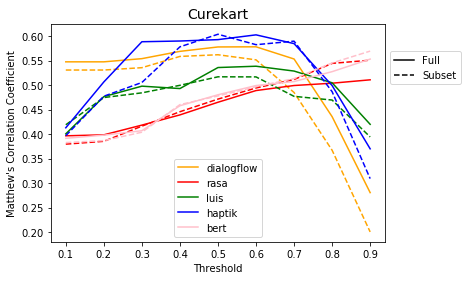}
\includegraphics[width=8cm, height=4.20cm]{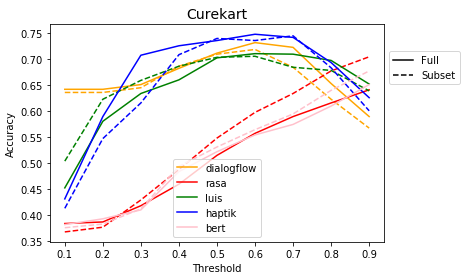}
\caption{Matthew's Correlation Coefficient and Accuracy across all datasets and platforms}
\centering
\label{fig:metrics}
\end{figure*}

\section{Datasets}
\label{sec:datasets}
We introduce HINT3, a collection of datasets shown in Table \ref{tab:data-stats} - \textbf{SOFMattress}, \textbf{Curekart} and \textbf{Powerplay11} each containing diverse set of intents in a single domain - mattress products retail, fitness supplements retail and online gaming respectively. Table \ref{tab:example-queries-hint3} shows few example intents of varying granularity in HINT3 dataset, along with examples of training queries created by domain experts and in-scope, out-of-scope queries received from real users.

\subsection{Training Data Collection}
Training data is prepared by a team of domain experts trying to emulate real users after in-depth research of historical user queries. The experts do not create an explicit set of out of scope queries primarily because the universe of such queries is infinitely big. Training datasets show class imbalance, occurrence of domain specific words, acronyms\footnote{\label{note1}github.com/hellohaptik/HINT3/tree/master/data\_exploration}. All training data queries are in English.

\subsubsection*{Dataset Variants}
In addition to \textbf{Full} training sets, we create \textbf{Subset} versions for each training set. For each class, after retaining the first query we iterate over the rest, discarding a query if it has an entailment score \citep{bowman-etal-2015-large} greater than 0.6 in both directions with any of the queries retained so far i.e. the subset version has the following property

\begin{equation*}
  \begin{aligned}
    E(x_a, x_b) \leq 0.6 \ \land \ E(x_b, x_a) \leq 0.6; \\
    a \neq b, a \in [1, |\hat{X}_{i}|], b \in [1, |\hat{X}_{i}|] \ \forall \ I
  \end{aligned}
\end{equation*}

where $I$ is the set of all intents, $\hat{X}_{i}$ is the set of queries retained for class $i$, $E(h, p)$ is the entailment scoring function with $h$ as hypothesis and $p$ as premise. We use ELMo model trained on SNLI \citep{peters-etal-2018-deep,parikh-etal-2016-decomposable} \footnote{https://demo.allennlp.org/textual-entailment} for $E(h, p)$. These are intended to evaluate performance with only semantically different sentences in the training set as ideally systems should already understand semantically similar queries to the ones present in the training set.

\begin{table*}[t]
\centering
\begin{tabular}{ccccccc}
\hline
 & \multicolumn{2}{c}{\textbf{SOFMattress}} & \multicolumn{2}{c}{\textbf{Curekart}} & \multicolumn{2}{c}{\textbf{Powerplay11}} \\
 & \textbf{Full} & \textbf{Subset} & \textbf{Full} & \textbf{Subset} & \textbf{Full} & \textbf{Subset} \\ \hline
\textbf{Dialogflow} & 73.1 & \textbf{65.3} & 75.0 & 71.2 & 59.6 & 55.6 \\
\textbf{RASA} & 69.2 & 56.2 & \textbf{84.0} & 80.5 & 49.0 & 38.5 \\
\textbf{LUIS} & 59.3 & 49.3 & 72.5 & 71.6 & 48.0 & 44.0 \\
\textbf{Haptik} & 72.2 & 64.0 & 80.3 & 79.8 & \textbf{66.5} & \textbf{59.2} \\
\textbf{BERT} & \textbf{73.5} & 57.1 & 83.6 & \textbf{82.3} & 58.5 & 53.0 \\ \hline
\end{tabular}
\caption{Inscope Accuracy at low threshold=0.1 for Full and Subset data variants}
\label{tab:in-scope-acc}
\end{table*}

\subsection{Test Data Collection and Annotation
}
Our test sets contain the first message received by live systems from real users over a period of 15 days. Inter-annotator agreement was 75.8\%, 80.0\% and 73.4\% for SOFMattress, Curekart and Powerplay11 respectively and conflicts were resolved by domain experts. One major reason for low inter-annotator agreement was unclear criteria for defining an intent which sometimes lead to overlapping intents of different levels of granularity, even after we had made sure to manually merge any conflicting or highly similar intents in the training data.

Directly coming from real users our test set queries also contain messaging slangs, acronyms, spelling mistakes, grammatical mistakes and usage of code-mixed languages\cref{note1}. Queries in non-Latin script or code-mixed languages were marked as out of scope (labelled as \texttt{NO\_NODES\_DETECTED}). Since live chat systems don’t cater all the queries related to a brand, our test set contains relevant out-of-scope queries received from users about that domain. Any identifiable information of users, brands was replaced with made-up values in both train and test sets.

\section{Benchmark Evaluation}

We evaluated the performance of our datasets on platforms like Dialogflow, LUIS, RASA and Haptik in addition to evaluating performance on BERT. All layers of BERT were fine-tuned with a learning rate of 4e-5 for up to 50 epochs with a warmup period of 0.1 and early stopping.

\subsection{Out-Of-Scope (OOS) prediction}

We use thresholds on the model's probability estimate for the task of predicting whether a query is OOS. We show performance on thresholds ranging from 0.1 to 0.9 at an interval of 0.1 to show the maximum performance a model can achieve irrespective of how we choose the threshold.

\subsection{Metrics}

We consider Accuracy and Matthew’s Correlation Coefficient\footnote{https://scikit-learn.org/stable/modules/model\_evaluation} as overall performance metrics for the systems. We use OOS recall \citep{larson-etal-2019-evaluation} to evaluate performance on OOS queries and accuracy of in-scope queries to evaluate performance on in-scope queries.

\section{Results}
\label{sec:results}


\begin{table*}[]
\centering
\resizebox{\textwidth}{!}{%
\begin{tabular}{@{}lllll@{}}
\toprule
\textbf{Test query} & \textbf{True label} & \textbf{Top predicted label} & \textbf{Sample training queries for True label} & \textbf{Sample training queries for predicted label} \\ \midrule
Ergo 7272 inches price? & MATTRESS\_COST & \textbf{L,H,D,R}: ERGO\_FEATURES & \begin{tabular}[c]{@{}l@{}}• Price of mattress\\ • Custom size cost\end{tabular} & \begin{tabular}[c]{@{}l@{}}• Features of Ergo mattress\\ • Tell me about SOF Ergo mattress\end{tabular} \\ \midrule
Trail option are there & 100\_NIGHT\_TRIAL\_OFFER & \begin{tabular}[c]{@{}l@{}}\textbf{L,H,D:} COD\\ \textbf{R:} EMI\end{tabular} & \begin{tabular}[c]{@{}l@{}}• Trial details\\ • How to enroll for trial\end{tabular} & \begin{tabular}[c]{@{}l@{}}• Can I get COD option?\\ • Can it deliver by COD\end{tabular} \\ \midrule
I require 75 inch 57 inch. Is it available? & SIZE\_CUSTOMIZATION & \begin{tabular}[c]{@{}l@{}}\textbf{L:} DISTRIBUTORS\\ \textbf{H,D,R:} WHAT\_SIZE\_TO\_ORDER\end{tabular} & \begin{tabular}[c]{@{}l@{}}• Will I get an option to Customise the size\\ • How can I order a custom sized mattress\end{tabular} & \begin{tabular}[c]{@{}l@{}}• Want to know the custom size chart\\ • Show me all available sizes\end{tabular} \\ \midrule
20 \% discount available on emi & OFFERS & \textbf{L,H,D,R:} EMI & \begin{tabular}[c]{@{}l@{}}• Want to know the discount\\ • Tell me about the latest offers\end{tabular} & \begin{tabular}[c]{@{}l@{}}• You guys provide EMI option?\\ • No cost EMI is available?\end{tabular} \\ \midrule
How will u deliver with this LockDown in place ? & NO\_NODES\_DETECTED & \textbf{L,H,D,R:} CHECK\_PINCODE & \multirow{2}{*}{-} & \multirow{2}{*}{\begin{tabular}[c]{@{}l@{}}• Do you deliver to my pincode\\ • Will you be able to deliver here\end{tabular}} \\ \cmidrule(r){1-3}
Covid19 how can you deliver & NO\_NODES\_DETECTED & \textbf{L,H,D,R:} CHECK\_PINCODE &  &  \\ \bottomrule
\end{tabular}%
}
\caption{Few examples of test queries in SOFMattress which failed on all platforms, \textbf{L}: LUIS, \textbf{H}: Haptik, \textbf{D}: Dialogflow, \textbf{R}: Rasa. NO\_NODES\_DETECTED is the out-of-scope label.}
\label{tab:platform-predictions}
\end{table*}

Figure \ref{fig:metrics} presents results for all systems, for both Full and Subset variations of the dataset. Best Accuracy on all the datasets is in the early 70s. Best MCC for the datasets varies from 0.4 to 0.6, suggesting the systems are far from perfectly understanding natural language.

In Table \ref{tab:in-scope-acc}, we consider in-scope accuracy at a very low threshold of 0.1, to see if false positives on OOS queries would not have mattered, what’s the maximum in-scope accuracy that current systems are able to achieve. Our results show that even with such a low threshold, the maximum in-scope accuracy which systems are able to achieve on Full Training set is pretty low, unlike the 90+ in-scope accuracies of these systems which have been reported on other public datasets like CLINC150 in \citep{larson-etal-2019-evaluation}. And, the in-scope accuracy is even worse for the Subset of the training data. 

\begin{table}[h]
\centering
{%
\begin{tabular}{cccc}
\hline
 & \textbf{\begin{tabular}[c]{@{}c@{}}SOF\\ Mattress\end{tabular}} & \textbf{Curekart} & \textbf{\begin{tabular}[c]{@{}c@{}}Power\\ play11\end{tabular}} \\ \hline
\textbf{Dialogflow} & \textbf{10.6} & 5.0 & \textbf{6.7} \\
\textbf{RASA} & 18.7 & 4.1 & 10.5 \\
\textbf{LUIS} & 16.8 & 1.2 & 8.3 \\
\textbf{Haptik} & 11.3 & \textbf{0.6} & 10.9 \\
\textbf{BERT} & 22.3 & 1.5 & 9.4 \\ \hline
\end{tabular}%
}
\caption{Percentage \textbf{drop} in Inscope Accuracy at low threshold=0.1 in Subset data as compared to Full}
\label{tab:score-drops}
\end{table}

Table \ref{tab:score-drops} shows percentage drop in in-scope accuracy on subset data across all systems as compared to in-scope accuracy on full data. The drop varies from 0.6\% to 22.3\% across datasets and platforms. In an ideal world, this drop should be close to 0 across all datasets, as if the system understands the meaning of queries in training data, its performance should not get affected at all by removing queries in training data which are semantically similar to the ones already present.

\begin{figure}[h]
\includegraphics[width=0.5\textwidth]{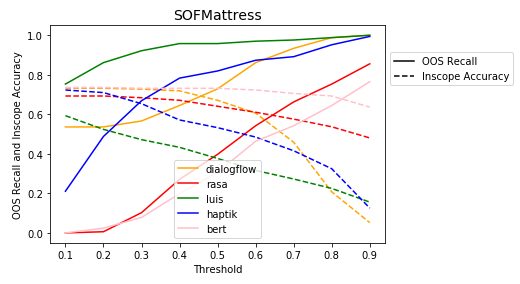}
\caption{Out-of-Scope (OOS) Recall at the cost of In-scope Accuracy for SOFMattress Full dataset}
\centering
\label{fig:sofmattress-oos}
\end{figure}

Analyzing few example queries which failed on all platforms in Table \ref{tab:platform-predictions} suggests that these models aren’t actually “understanding” language or capturing “meaning”, instead capturing spurious patterns in training data, as was also pointed in \citep{bender-koller-2020-climbing}. Predicting based on these spurious patterns, which models latch on to during training, leads to models having high confidence even on OOS queries. Figure \ref{fig:sofmattress-oos} shows this behaviour on SOFMattress Full dataset, as significant percentage of OOS queries have high confidence scores on all systems, except LUIS, for which it is at the cost of in-scope accuracy.

\section{Conclusion}
\label{sec:conclusion}

This paper analyzed intent detection on 3 new datasets consisting of both in-scope and out-of-scope queries received on 3 live chat bots over a period of 15 days. Our findings indicate that there’s a significant gap in performance on crowd-sourced datasets vs in a real world setup. NLU systems don’t seem to be actually “understanding” language or capturing “meaning”. We believe our analysis and dataset will lead to developing better, more robust dialogue systems.

\section*{Acknowledgments}

We are grateful to Bot Analysts at Haptik, especially Aaron Dsouza\footnote{Reachable at aaron.dsouza@haptik.ai}, who helped us open-source HINT3 datasets. We also want to thank clients of Haptik who allowed us to share queries received on their bots with the research community.

\bibliographystyle{acl_natbib}
\bibliography{emnlp2020}
\end{document}